# SLEEPNET: Automated Sleep Staging System via Deep Learning

**Siddharth Biswal** SBISWAL7@GATECH.EDU
*Georgia Institute of Technology*

**Joshua Kulas** JKULAS3@GATECH.EDU
*Georgia Institute of Technology*

**Haoqi Sun** HSUN004@E.NTU.EDU.SG
*Nanyang Technological University*

**Balaji Goparaju** BGOPARAJU@MGH.HARVARD.EDU
*Massachusetts General Hospital*

**M Brandon Westover** MWESTOVER@MGH.HARVARD.EDU
*Massachusetts General Hospital*

**Matt T Bianchi** MTBIANCHI@MGH.HARVARD.EDU
*Massachusetts General Hospital*

**Jimeng Sun** JSUN@CC.GATECH.EDU
*Georgia Institute of Technology*

## Abstract

Sleep disorders, such as sleep apnea, parasomnias, and hypersomnia, affect 50-70 million adults in the United States (Hillman et al., 2006). Overnight polysomnography (PSG), including brain monitoring using electroencephalography (EEG), is a central component of the diagnostic evaluation for sleep disorders. While PSG is conventionally performed by trained technologists, the recent rise of powerful neural network learning algorithms combined with large physiological datasets offers the possibility of automation, potentially making expert-level sleep analysis more widely available.

We propose `SLEEPNET` (Sleep EEG neural network), a deployed annotation tool for sleep staging. `SLEEPNET` uses a deep recurrent neural network trained on the largest sleep physiology database assembled to date, consisting of PSGs from over 10,000 patients from the Massachusetts General Hospital (MGH) Sleep Laboratory.

`SLEEPNET` achieves human-level annotation performance on an independent test set of 1,000 EEGs, with an average accuracy of 85.76% and algorithm-expert inter-rater agreement (IRA) of $\kappa = 79.46\%$, comparable to expert-expert IRA.

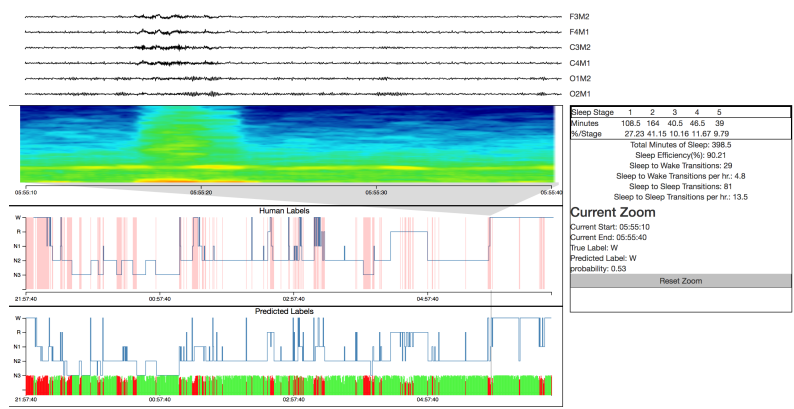

Figure 1: Web interface of `SLEEPNET`. In top-panel 1 and 2 we plot the raw EEG data the corresponding spectrogram. Below them, we show the ground truth (staging performed by a clinical expert) and `SLEEPNET` predicted labels and algorithm's confidence (red means low confidence while green means high confidence). On the right hand-side is a panel showing sleep related summary statistics.

# 1. Introduction

Sleep disorders including sleep apnea are a significant health concern in the US and throughout the world. An estimated 50-70 million people in the United States currently suffer from sleep related disorders(Hillman et al., 2006). The central diagnostic tool in the evaluation of sleep apnea is the overnight sleep study, during which several physiological signals are recorded, including the electroencephalogram (EEG); the overall recording is termed as polysomnogram (PSG). A central part of analyzing an overnight PSG is characterizing the patient's sleeping pattern based upon visual scoring of the EEG.

Manual scoring carries limitations of inter-rater reliability and time demands of sleep technologists. An 8-hour PSG study can take over one hour of technologist's time to generate complete sleep report. Much of that time is spent on manual annotation. Automation could alleviate those manual efforts if performance was sufficiently accurate to assist or even supplant manual staging. A major challenge in automation is the vast heterogeneity observed in EEG time series data obtained during PSG. Theoretically, two approaches could be attempted to overcome this heterogeneity: either create multiple algorithms and apply them to clinically defined sub-populations (by age, sex, sleep disorder, medication, etc) expected to have distinct EEG patterns, or train a single powerful algorithm on a large data set of PSGs that captures the heterogeneity across patients while still maintaining performance accuracy. We undertook the latter challenge using a training and testing set of over 10K subjects, chosen without regard to age, sex, sleep disorder, or treatment modality (3.2 TB raw EEG data). In this way, the clinical dataset in this work is large and heterogeneous, two orders of magnitudes larger than in previous studies.

In recent years deep neural networks have achieved impressive performance in several challenging real-world tasks that were previously the exclusive domain of human experts including the ones in medical fields Esteva et al. (2017); Gulshan et al. (2016). Availability of large amounts of data to learn complex data distributions is a key reason for the success of neural networks. Most published work to date comprise proof of concept research projects and limited effort has been devoted to actual deployment of deep neural networks into clinical practice.

In this paper, we propose a specialized clinical decision support tool, `SLEEPNET`, for automated sleep-wake staging. `SLEEPNET` applies deep neural network modeling on EEG signal features derived from routine clinical PSG data. The deployed algorithm presents the algorithm's inferences along with the raw EEG data and automatically calculates sleep quality statistics in an interactive web interface for clinicians and researchers. We experiment with many different variations of the algorithm and input features. The best performing instance of `SLEEPNET` uses expert-defined features to represent each 30-sec interval and learns to annotate EEG using a recurrent neural network (RNN). To measure performance of `SLEEPNET`, we compare automated staging results with the de facto gold standard defined by sleep technologists' manual annotations. Inter-rater agreement (IRA) for sleep staging (in terms of Cohen's Kappa) between human experts is approximately 65-75%(Danker-hopfe et al., 2009). In this study, SLEEPNET achieves an expert-algorithm Kappa value of 79%, with an overall accuracy of 86%, comparable to human-level scoring performance.

The main contributions of this work are as follows:

- We propose an end-to-end deep neural network, `SLEEPNET`, that learns to annotate sleep stages based on a large amount of PSG data. This is a novel deep neural network which can be extended to other types of EEG annotation and classification tasks.

- We have evaluated our proposed `SLEEPNET` on 10,000 overnight PSGs containing about eight hours of data per patient (9,000 for training and 1,000 for testing). This is so far the largest evaluation on an EEG classification problem.

- We have deployed `SLEEPNET` at Massachusetts General Hospital (MGH) and have performed initial quantitative and qualitative evaluations of the deployed model, which confirm the clinical and research value of the proposed system.

After discussing the previous approaches to EEG classification and deep neural networks in Section 2, we describe training and deployment of our proposed SLEEPNET system in Section 3. We present experiments evaluating the performance of SLEEPNET with both qualitative and quantitative evaluation in Section 4. In Section 5, we also present case studies of SLEEPNET deployed in two clinical laboratories at MGH and conclude the work with discussion on future directions in Section 6.

## 2. Related work

**EEG classification:** Much of the work on EEG classification has focused on the problems of seizure prediction and detection, and brain computer interfaces (Mirowski et al., 2008; Shoeb and Guttag, 2010). These efforts have employed a wide variety of machine learning methods (Subasi, 2007; Garrett et al., 2003; Mormann et al., 2007). Emphasis on feature engineering guided by physiological insights is a common theme in this work.

Machine learning approaches for sleep stage annotation in EEG signals have been also previously proposed (Berthomier et al., 2007; Anderer et al., 2010; Fraiwan et al., 2010). Some of these have applied neural networks (Schaltenbrand et al., 1996). Commercial software has also been developed for automatic sleep stage scoring (e.g. ZMachine®) and evaluated in previous studies (Wang et al., 2015). However, most prior studies involve small numbers of subjects, usually <100. The largest study to date involved 590 recordings, thus these studies may not adequately address variability between subjects (Anderer et al., 2005). The small study datasets may not contain sufficient variety to ensure generalization to broader populations.

**Deep neural networks for healthcare applications:** Deep neural networks have demonstrated performance gains in many challenging tasks such as image classification, speech recognition, image captioning (Krizhevsky et al., 2012; Hannun et al., 2014; Karpathy et al., 2014). Two popular deep learning models, namely convolutional neural networks (CNN) and Recurrent neural networks(RNN), have achieved considerable success in health applications. Among CNN variants, AlexNet was one of the first successful deep CNN to successfully classify 1000 different categories of images (Krizhevsky et al., 2012). Over time new architectures such as InceptionNet and ResNet have been proposed which use different key methods such as residual connections and multiple convolutions to the same input (Szegedy et al., 2015; He et al., 2015). Various CNN models have been utilized for analyzing medical images to achieve human level performance for diagnosis tasks (Gulshan et al., 2016; Esteva et al., 2017).

RNN is another powerful deep learning model specialized for sequential data such as continuous time series and sequences of discrete events. RNNs have had great success in speech recognition, handwriting recognition, and machine translation (Sutskever et al., 2014; Graves, 2013; Graves et al., 2008). In health care applications, RNNs have also demonstrated success on predictive modeling problems using electronic health records (Choi et al., 2016a,b; Lipton et al., 2015).

Following the recent development of deep neural networks, methods have been proposed to learn feature representation from EEG data. Recently, a method was proposed to learn an EEG representation by converting the signal into an image using the location of electrodes and applying deep a CNN to the image(Bashivan et al., 2015). Convolutional neural networks have also been applied to hand-chosen features for epileptic seizure recognition(Mirowski et al., 2008). Compact CNNs have been proposed to learn representation of EEG for brain computer interface tasks (Lawhern et al., 2016). These successful applications to EEG data suggest that deep learning methods have potential for analyzing EEG data from PSGs to extract efficient representations for automatic sleep-wake stage annotation.

## 3. SLEEPNET System Description

In this section we present the overall system architecture of SLEEPNET in details.

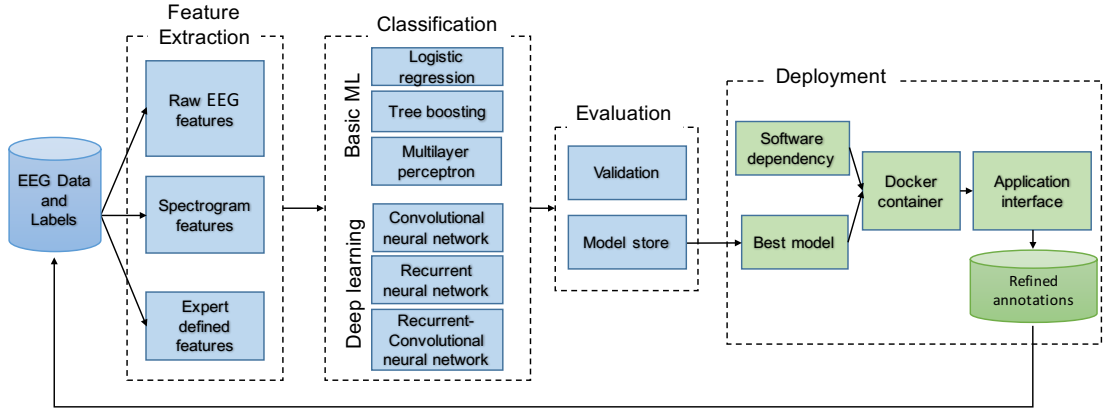

(a) Analytic pipeline of `SLEEPNET`. The blue color components correspond to model training module. The green color components belong to the model deployment module.

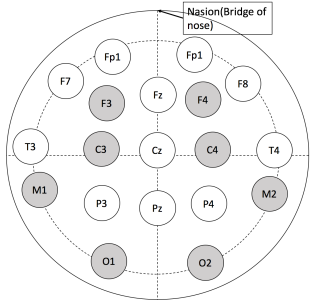

(b) Electrode locations of EEG recording

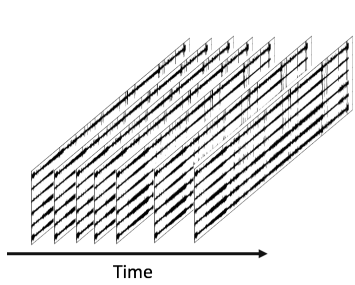

(c) EEG raw feature representation

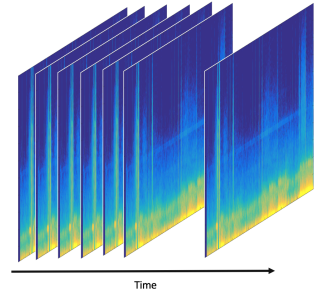

(d) Spectrogram representation

Figure 2: System architecture of `SLEEPNET`

## 3.1 System overview

`SLEEPNET` consists of a *training* module and a *deployment* module. The system architecture is depicted in Figure 2a.

**Training module:** The objective of the model training module is to find an accurate annotation algorithm that can take the multi-channel EEG data as input and automatically output a sequence of sleep stages, with one stage label assigned to each 30 second epoch. This sequence of stages is conventionally called a *hypnogram*. To generate such an annotation algorithm, the model training module extract features from the EEG data and then find classification algorithms to identify the best feature and algorithm configuration. As we describe in more detail next, we find that the best configuration among those explored is the combination of expert-defined features and a recurrent neural network (RNN) model. Model training is described in Sections 3.2 through 3.6.

**Deployment module:** The best performing model is deployed via a light-weight docker image at the point of care at MGH via a web application. The deployment module can load new PSG files generated from a clinic (e.g., sleep lab or neuro ICU) and automatically assign a sequence of sleep stages. A web interface is also developed to display the raw EEG data as well as the spectrogram and some key summary statistics based on the computed hypnogram. If human-assigned labels are available (e.g., labels created by technicians in the sleep lab), `SLEEPNET` displays both the human labels and algorithm predicted labels together and highlight regions of disagreement. Details of the model deployment is described in Section 3.7.

### 3.2 Description of EEG data and Notations

EEG signals consist of local voltage potentials recorded over a brain region. Usually EEG signals are collected from electrodes placed over multiple regions of the head, represented as different channels in the EEG data. The length of EEG data collected varies depending on the clinical task. EEG data in PSG consists of data from 6 different channels, i.e. F3, F4, C3, C4, O1 and O2, each referenced to the contralateral mastoid indicated by M1 or M2(Böcker et al., 1994). For example, F3-M2 and F4-M1 represent two frontal EEG channels. In the Figure 2b., we show a schematic of the locations of the electrodes.

For each 30 second epoch, the sleep EEG signals were annotated as belonging to one of five stages: wake(W), rapid eye movement(R), non-REM stage 1 (N1), non-REM stage 2 (N2), or non-REM stage 3(N3). All annotations of 10,000 PSG studies were performed by certified sleep technologists according to American Academy of Sleep Medicine (AASM) standards (Iber et al., 2007).

Next we introduce some necessary notation: EEG signals for each patient are represented as a matrix $\mathbf{X}_t \in \mathbb{R}^{N_i \times 6}$ where $N_i$ is the number of samples from 6 different channels of EEG for the $i'th$ patient. Sleep stages are represented as a sequence $\mathbf{y} \in \mathbb{R}^{n_i}$ where $n_i$ is the number of 30 seconds windows in the entire recording, and $y \in N1, N2, N3, R, W$. Since the EEG signals are sampled at 200Hz, $n_i$ and $N_i$ have the following relationship: $n_i = \frac{N_i}{200 \times 30}$. The classification task in model training is to infer a hypnogram $\hat{\mathbf{y}} \in \mathbb{R}^{n_i}$ that is close to the ground-truth hypnogram $y$ (i.e. a hypnogram that an expert would produce by conventional manual scoring of the EEG), such that misclassification between $\hat{\mathbf{y}}$ and $\mathbf{y}$ is small.

Table 1: Basic Notations

| Symbol | Definition | Dimensionality |
|---|---|---|
| $\mathbf{X}_t$ | raw EEG time series | $N_i \times 6$ |
| $\mathbf{X}_f$ | EEG into frequency domain | $n_i$ |
| $\mathcal{F}_R$ | Raw EEG tensor feature | $6000 \times 6 \times n_i$ |
| $\mathcal{F}_S$ | Spectrogram tensor feature | $29 \times 257 \times n_i$ |
| $\mathbf{F}_E$ | Expert defined feature | $96 \times n_i$ |

For simplicity every notation is for a single patient

### 3.3 Feature Extraction

Now we describe different types feature representations for EEG signals which are computed in the SLEEPNET model training module. For simplicity, we describe each feature only for a single patient.

#### 3.3.1 RAW EEG:

We consider the EEG data $\mathbf{X}_t$ for each 30-second epoch to create this raw EEG feature representation. This is represented as a 3-way tensor $\mathcal{F}_R \in \mathbb{R}^{6000 \times 6 \times n_i}$ which includes 6000 samples from 6 different channels from the 30-second EEG data. Figure 2c shows a visualization of the raw EEG feature as a 3-way tensor.

#### 3.3.2 SPECTROGRAM:

The EEG signal $\mathbf{X}_t$ can be represented in the frequency domain using the Fourier transform. The frequency domain representation is denoted as $\mathbf{X}_f$ and is obtained by Fast Fourier Transform(FFT). To perform robust power spectrum estimation, we have used multitaper spectral analysis (MTSA) (Thomson, 1982). The spectrogram is a way of representing the frequency content of the EEG over time. Each 30-second epoch was segmented into 29 sub-epochs of that are 2 seconds long with 1-second overlap. For each sub-epoch, we used MTSA to estimate power spectral density in 257 frequency bins from 0-100Hz. Further, the resulting spectrogram feature is denoted as another 3-way tensor $\mathcal{F}_S \in \mathbb{R}^{29 \times 257 \times n_i}$, where each spectrogram has dimension $29 \times 257$ and $n_i$ is the number of 30-second windows in the entire EEG signal. Figure 2d shows a 3-way tensor of spectrogram for one 30-second EEG epoch.

### 3.3.3 EXPERT DEFINED FEATURES:

We extract various time domain and frequency domain features from each 30-second epoch of EEG data. In Table 2, we list the expert defined features extracted from both time and frequency domains. We also extracted two different types of time domain features. Line length is included as a measure of amplitude and frequency oscillations in the EEG(Esteller et al., 2001). Kurtosis is included to measure the presence of extreme values(Zoubek et al., 2007).

In order to extract frequency domain features, we first segment each 30-second epoch into 29 sub-epochs of 2 seconds long with 1-second overlap and estimate the power spectral density of each sub-epoch. This results in 6 spectrograms from 6 channels with $29 \times 257$ dimensions. To reduce noise in spectral features we averaged spectrograms from contralateral channels, i.e. F3-M2 and F4-M1, C3-M2 and C4-M1, O1-M2 and O2-M1. which reduces the dimension to 3 averaged spectrogram of $29 \times 257$ dimension. Then we find indices for each of three frequency bands: delta (0.5-4Hz), theta (4-8Hz), alpha (8-12Hz) in the spectrogram matrix. We normalize these power estimates by the total power from 0-20Hz. From the resulting matrix, we obtain 95th percentile (robust version of maximum), minimum, mean and standard deviation. We also include features reflecting relative power such as i.e. delta/theta, delta/alpha and theta/alpha ratios. The kurtosis of the spectrogram in the delta, theta, alpha and sigma (12-20Hz) bands were also extracted to capture transient bursts such as sleep spindles.

The final dimension of the expert defined feature vectors are $\mathbf{F}_E \in \mathbb{R}^{96 \times n_i}$.

Table 2: Expert Defined Feature List

| Domain | Feature | #features |
|---|---|---|
| Time Domain | Line length | 6 |
| | Kurtosis | 6 |
| Frequency Domain | delta-total power ratio | 12 |
| | theta-total power ratio | 12 |
| | alpha-total power ratio | 12 |
| | delta-theta power ratio | 12 |
| | theta-alpha power ratio | 12 |
| | delta-alpha power ratio | 12 |
| | Kurtosis of delta band of spectrogram | 3 |
| | Kurtosis of theta band of spectrogram | 3 |
| | Kurtosis of alpha band of spectrogram | 3 |
| | Kurtosis of sigma band of spectrogram | 3 |

### 3.4 Classification in SLEEPNET

Once the features are constructed, the model training module compares various classification models including basic machine learning methods such as logistic regression, tree boosting, multilayer perceptrons and deep learning methods. Next, we describe the deep learning methods in more detail.

### 3.4.1 CONVOLUTIONAL NEURAL NETWORK:

Convolutional Neural Networks (CNNs), introduced by Le Cun et al.(LeCun et al., 1989) comprise a class of neural network architecture which use a series of convolutional filters, non-linear activation functions and pooling layers to minimize a loss function. Given a feature matrix $\mathbf{X}$, the convolutional layer convolves **X** with k filters $\{W_i\}_k$ to produce preactivation maps **H**.

$$H_i = W_i * X + b_i, i = 1, ...., k \qquad (1)$$

where the symbol $*$ denotes the convolution operation and bias $b_i$ is a bias parameter. In `SLEEPNET`, we use Rectifier Linear Unit(ReLu) as non-linear activation functions(Glorot et al., 2011). Given a simple feature map $H_i$, a ReLu function is defined as $\hat{H_i} = max(0, H_i)$.

After the non-linear activation unit, we pass the features through a max-pooling layer to reduce the spatial size of the representation. Figure 3 shows a diagram of the CNN architecture used in our experiments. Our final model had stacked 32 $3 \times 3$ filters, 64 $3 \times 3$ filters, 128 $3 \times 3$ filters each followed by max pooling layer and finally connected to a fully connected layer. We also tried other filter sizes ($5 \times 5$ and $7 \times 7$) but $3 \times 3$ filter gives the best test-set performance.

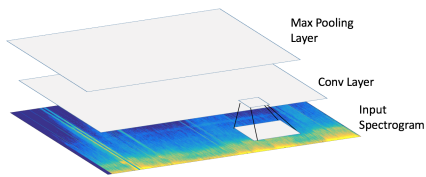

Figure 3: CNN with Spectrogram as input

In our experiments, we have used a 2D CNN for processing spectrograms which are of $29 \times 257$ dimension. The raw waveform from 6 channels were averaged to a single channel and passed to a 1D CNN for classification.

### 3.4.2 RECURRENT NEURAL NETWORKS:

RNNs are tailored toward modeling sequential data. Figure 4 illustrates the process of feeding RNNs with input features with every timestamp corresponding to a 30-second epoch. More specifically, we provide feature representation of EEG up to the current timestamp $t$, to classify the feature representation at timestamp $t$. At timestamp $t$, nodes get input from the current time step $x_t$ and from previous hidden states $h_{t-1}$ to provide output $y_t$. The following equation specifies the simple formulation of an RNN.

$$y_t = \text{Softmax}(W_{yh}h_t + b_y) \qquad (2)$$
$$h_t = \sigma(W_{hx}x_t + W_{hh}h_{t-1} + b_h) \qquad (3)$$

where $W_{hh}$ is the weight matrix associated with between hidden layers, $W_{hx}$ is weight matrix associated with between hidden layers and input layer and $W_{hy}$ is the weight matrix associated between hidden layer and output layer. $\sigma$ is the activation function such as sigmoid or tanh. The vectors $b_h$ and $b_y$ are bias parameters associated with hidden layer and output layer. We implemented the RNN formulation using Long Short Term Memory (LSTM) in Tensorflow(Graves, 2013; Hochreiter and Schmidhuber, 1997).

In order to incorporate regularization to avoid overfitting, we have added dropout(Srivastava et al., 2014) to our multilayer RNN architecture. In our experiments, we have experimented with different numbers of layers, activation functions, and dropout probability to optimize the structure of the RNN. Finally, our optimal model had 5 layers of LSTM cells with tanh activation function and dropout keep probability of 0.9. The size of the hidden layer was set to 1000.

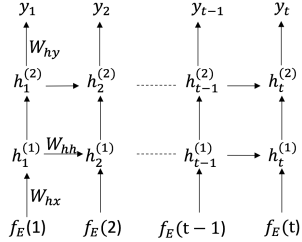

Figure 4: Multi-layer RNN for SLEEPNET

### 3.4.3 RECURRENT-CONVOLUTIONAL NEURAL NETWORKS:

By combining a RNN with CNN, we can have a hybrid model, namely, Recurrent-Convolutional Neural Networks (RCNN), which is able to extract features present in a spectrogram and preserve the long-term temporal relationship present in the EEG data.

In RCNN model, a CNN first processes the spectrogram over all non-overlapping 30-second windows of EEG data to learn feature representation. Here CNN extracts the "spatial" features from EEG which is time invariant and independent at each step. That feature representation is passed to a RNN model (LSTM in this case), which learns the temporal dependency present of the spatial feature extracted by CNN. More specifically, RCNN model passes each 30-second spectrogram $\mathcal{F}_S$ to a CNN model to produce a fixed-length spatial feature representation $\phi_\mathbf{t}^\mathbf{i}$.

$$\phi_\mathbf{t} = \text{CNN}(\mathcal{F}_S) \qquad (4)$$

where $t = 1, 2, \ldots n_i$. $\phi_\mathbf{t}$ is 4,352 dimensional vector after a $29 \times 257$ spectrogram is passed through a CNN. The resulting spatial features over time become a sequence denoted by $\phi_\mathbf{1}, \phi_\mathbf{2} \ldots \phi_\mathbf{t}$. Taking the above sequence as input, RNN maps an current input $\phi_\mathbf{t}$ and previous time-step hidden state $\mathbf{h_{t-1}}$ to the next hidden state $\mathbf{h_t}$. Then the hidden state $\mathbf{h_t}$ pass to a softmax function to generate the final sleep stage prediction $\mathbf{y}$ described in Eq 2. We use categorical cross-entropy as the loss function of the classification step from spectrogram sequence to sleep stage. The entire model (including CNN and RNN) are trained together via backpropagation. We again use the Long Short Term Memory(LSTM) model as the implementation for the RNN (Graves, 2013).

### 3.5 Evaluation

The evaluation is done by splitting the data into train, validation and test set. And the best model is persisted into the model store for deployment. For all classification methods, we use EEG from 1K patients as the final test set. The remaining 9K patients are used for training and validation. For deep learning model, we use 300 patients as validation. Note that each patient corresponds to 8 hours' data about 950-1000 labels (one for each 30-second epoch).

We performed 50 iterations of random search over a set of parameter choices for hyper-parameter tuning. Here are the parameter choices:

- Learning rate : [0.01, 0.001, 0.001, 0.0001, 0.00001]
- Look back steps in RNN : [3,5,10,20,30]
- Dropout rate : [0.0, 0.2,0.4,0.6, 0.8, 0.9]
- Number of hidden units in RNN : [100, 200, 400, 800, 1000,2000, 5000]
- Number of layers in RNN: [1,2,3,5,7,8,15]
- Filter Size in CNN : [$(3 \times 3), (5 \times 5), (7 \times 7)$]

We also experimented with normalization methods such as Batch normalization.(Ioffe and Szegedy, 2015).

### 3.6 Implementation Details of Model Training Module

We implemented deep learning models of SLEEPNET with Tensorflow 0.12.1 (Team, 2015). For training models, we used a machine equipped with Intel Xeon E5-2640, 256GB RAM, four Nvidia Titan X and CUDA 8.0. We have used scikit-learn for logistic regression, Multi-layer perceptron (Pedregosa et al., 2011). We have used XGBoost which is an efficient implementation of Tree Boosting algorithm (Chen and Guestrin, 2016).

### 3.7 Model Deployment Module

Now we describe the preliminary deployment of SLEEPNET at Massachusetts General Hospital(MGH). All qualitative evaluations are performed by physicians (MTB, MBW).

#### 3.7.1 Deployment Architecture

There are significant infrastructure differences between training and deployment environment. The model training phase was performed on a server with high performance GPU cards (NVIDIA Tesla Pascal) to enable efficient iteration over model choices. To deploy the model at a commodity laptop or desktop at the point of care, we use Docker technology to ease the deployment effort. Using Docker, we package our trained model with all the required parts as an image. We use the same Docker image in the deployment environment at MGH to run an instance of the image.

#### 3.7.2 Deployment Process at MGH

We developed a web application which runs the trained model in the backend using a Docker container to score sleep EEG cases. To make software usage easy, clinicians just have to select a case, after which the application runs the trained model in the backend and shows a report page with clean visualization.

A screenshot of the web application and visualization is shown in 1. The visualization has 4 sub-panels. The first shows the EEG waveform, the 2nd and 3rd panels show the expert-assigned sleep stages and the stage labels predicted by the algorithm. The 4th sub-panel shows an average spectrogram from 6 EEG channels. In the figure, a clinician has selected a particular 30-second epoch for closer inspection. The raw EEG signal and spectrogram are shown as zoomed-in views.

SLEEPNET automatically calculates and displays standard summary statistics for sleep EEG, including the number of minutes spent in each stage, sleep efficiency (time spent in sleep / total recording time, i.e. (N1+N2+N3+R)/( N1+N2+N3+R+W), and sleep fragmentation indices. These statistics provide physicians with a quick summary of a patient's sleep quality.

### 4. Experiments

#### 4.1 Dataset Description

This dataset was collected at Massachusetts General Hospital Sleep Laboratory. This study was approved by Partners Institutional Review Board for retrospective analysis. EEG signals are usually collected with respect a reference electrode or chosen as the zero level. In our dataset EEG signals were referenced to electrodes located M2 or M1 position. The EEG signals were sampled at 200 Hz from F3-M2 and F4-M1, C3-M2 and C4-M1, O1-M2 and O2-M1 channels each referenced to the contralateral mastoid(M2 or M1)(Böcker et al., 1994). We included EEGs if three or more sleep stages were present.

EEG signals were annotated by PSG technicians in non-overlapping 30-second epochs according to AASM standards(Iber et al., 2007). Each epoch was labeled by a single experienced PSG technologist as one of five stages: wake (W), rapid eye movement (REM), Non-REM stage 1 (N1), Non-REM stage 2 (N2) and Non-REM stage 3 (N3). Table 3 provides the summary statistics.

Table 3: Descriptive statistics of dataset

| Dataset Property | Number |
|---|---|
| Number of Patients | 10,000 |
| Hours of EEG data | 80,000 |
| Raw data storage | 3.2 TB |

We have represented EEG data using 3 different types of feature representation.

- **Raw waveform($\mathcal{F}_R$):** In this representation, we feed the raw waveform from 6 channels of each 30-second EEG data segment to the models as input. The data dimensions for a 30-second window are $6 \times 6000$.

- **Spectrogram ($\mathcal{F}_S$):** Averaged spectrogram from multiple channels of EEG is provided as the input feature to different models. Spectrogram dimensions are $29 \times 257$.

- **Expert-defined features($\mathbf{F}_E$):** Expert-defined features are fed as the input to different models. The dimension of the feature was 96 for a 30 second epoch of sleep EEG. The detailed feature definitions are presented in Section 3.3.3.

**Classification Methods:** We conduct evaluation using the following classification algorithms.

- **Logistic Regression(LR):** We train a logistic Regression(LR) model using input features from each 30-second epoch of EEG data and the sleep stage as the label.

- **Tree Boosting(TB):** We use the same setup as logistic regression model in a Tree boosting classifier (Chen and Guestrin, 2016).

- **MultiLayer Perceptron (MLP):** We train a MLP with the setup as logistic regression. We used cross entropy loss function and trained using backpropagation. The hidden layer had 800 units.

- **Convolutional Neural Network(CNN):** We train a CNN with 2D convolutions to accept feature vector in terms of expert defined features or spectrogram of a 30-second epoch of sleep EEG data. In our experiments we have evaluated the effect of depth and different non-linearity options for building a deep CNN. We used a CNN with 1D convolutional layers to process raw EEG features. The details of the implmentation is presented in section 3.4.1.

- **Recurrent Neural Network(RNN) :** We have also trained a multi-layered RNN with LSTM cells to process sequence of features to produce sleep stage classes corresponding to the sequence.

- **Recurrent-Convolutional Neural Network(RCNN):** In this model, we process the spectrograms using a convolutional neural network. The activation maps of the final layer of the CNN are passed to the 2-layer RNN for sleep stage classification.

### 4.2 Accurate Model Classification

We have chosen accuracy and Cohen's kappa as performance metrics to evaluate of different combinations of feature representation and models. Cohen's kappa is a common metric for sleep study, which measures agreement between raters each rating N items into C mutually exclusive categories. The formula for Cohen's kappa is $\kappa = \frac{p_0 - p_e}{1 - p_e}$ where $p_o$ is the relative agreement between raters and $p_e$ is the agreement probability by a random chance. In table 4, we show our results for different combinations of features and methods. Multilayer RNN with expert-defined features has the best performance, although RCNN also has very competitive performance. Overall, all the deep learning models consistently outperform the traditional classification methods such as logistic regression, tree boosting and MLP, which confirmed the power of deep learning on modeling EEG data. Note that the performance of Kappa 79.46% from RNN is significantly higher than many prior studies(Anderer et al., 2005). More impressively, `SLEEPNET` actually performed well on a much larger and broader cohort of 10K patients, which is significantly larger than the prior works.

|       | Expert Defined Features |       | Spectrogram Features |       | Waveform Features |       |
|-------|:---:|:---:|:---:|:---:|:---:|:---:|
| Model | Accuracy | Kappa | Accuracy | Kappa | Accuracy | Kappa |
| LR    | 68.54 | 63.88 | 66.54 | 66.61 | 67.43 | 62.71 |
| TB    | 75.67 | 69.47 | 71.61 | 65.37 | 72.36 | 66.37 |
| MLP   | 72.23 | 68.41 | 70.23 | 66.71 | 69.56 | 64.21 |
| CNN   | –     | –     | 77.83 | 71.45 | 77.31 | 71.47 |
| RNN   | **85.76** | **79.46** | 79.21 | 73.83 | 79.46 | 72.46 |
| RCNN  | 81.67 | 76.38 | 81.47 | 74.37 | 79.81 | 73.52 |

Table 4: Performance of different feature representations with model combinations

We also show the confusion matrix to illustrate the performance of the best performing RNN model in the figure 5. In the five different stages of sleep, N1 is hardest to classify, which aligns with clinical expectation.

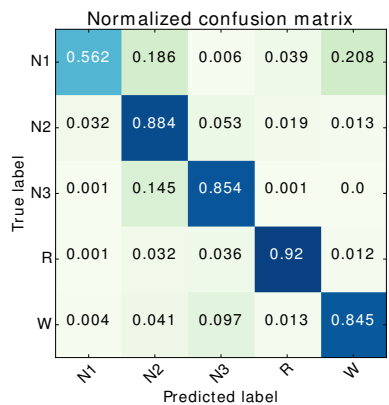

Figure 5: Normalized Confusion matrix for the performance of Expert defined features with RNN

### 4.3 Efficient Model Training and Model Scoring

We performed extensive experiments to evaluate the training and scoring time of different model and feature representation combinations. Figure 6 shows the training time for different models over 9000 patients. This shows that more complex models with raw data input takes longer as the dimensionality of feature space is larger.

We also compared scoring time different models for the entire 8-hour EEG recording of a patient (including feature extraction time) in Figure 7. The overall time spent varies from 2 to 5 minutes, which is sufficiently small to satisfy the deployment requirement. Note that the annotation results can be progressively displayed as the scoring process applies to each 30-second epoch of EEG data.

### 4.4 Sensitivity Analysis of Model Training

**Long-term temporal dependency:** We evaluate model performance with different look-back steps in RNN. Figure 8a shows that performance of the model increases as the number of lookback steps. This indicates that long-term temporal dependency does help improving sleep stage classification.

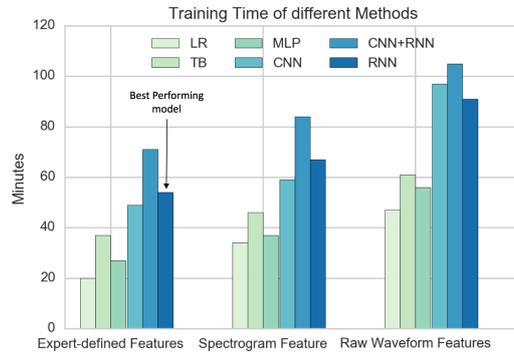

Figure 6: Training time comparison of different feature and model combinations for training 9000 patients

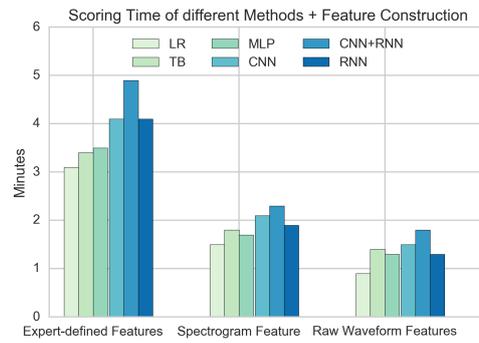

Figure 7: Scoring time comparison of single subject of different method with Feature Extraction

**Varying the number of training patients:** Figure 8b shows that Kappa on the test set continuously increases as the number of training subjects.

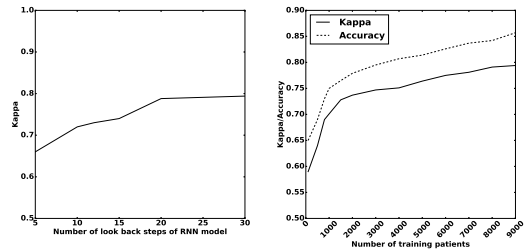

(a) Varying number of lookback steps  (b) Varying number of training patients

Figure 8: Impact of Data Parameters

# 5. Deployment Case Studies

We conduct two case studies on two very different clinical environments, namely, 1) **Sleep lab** where the technicians already have the expertise to annotate EEG data for sleep stage and look for tools to help speedup their annotation process and ensure better quality; 2) **Neurology ICU** where patients (such as epilepsy patients) do have continuous EEG data while the clinician may not have the capability or time to provide high quality sleep stage annotation.

## 5.1 Qualitative evaluation in MGH Sleep Lab

A sleep clinician (MTB) evaluated the system on 10 overnight EEG recordings conducted in the MGH Sleep Laboratory and provided the following detailed feedback. Loading the data and using the interface are each fast and conform with rapid workflow typical of clinical PSG interpretation. Visualization is natural, and similar to viewers used in clinical flow that allows clicking within a hypnogram representation of the full night to see more detailed physiology. Using this viewer, one can easily recognize and attend to discrepancies between technologist scoring and algorithm scoring, complemented by visualization of algorithm confidence in the class output. The overall feedback is extremely positive and `SLEEPNET` achieved the core design objectives.

## 5.2 Qualitative evaluation in MGH Neurology Intensive Care Unit (ICU)

A clinician specializing in critical care brain monitoring (MBW) also evaluated the system on 10 continuous EEG recordings of >24 hours duration each from ICU patients undergoing EEG monitoring on the MGH Critical Care EEG Monitoring Service (CCEMS), and provided the following feedback. Sleep deprivation is among the most common complaints among survivors of critical illness, yet most ICUs lack the ability to measure sleep except in specialized research studies(Novaes et al., 1999; Nelson et al., 2002). Whereas clinical services by CCEMS provide EEG monitoring for the purpose of detecting seizures as a complication of critical illness, sleep analysis is too time intensive and requires expertise outside the scope of CCEMS. In this pilot study of deployment, SLEEPNET allowed us to rapidly characterize sleep in individual ICU patients without adding significant clinical work. We were able to confirm findings from previous research on sleep in the ICU showing that sleep is often severely disrupted(Gabor et al., 2001; Harrell and Othmer, 1987). In these 10 patients, sleep quality was poor, characterized by low efficiency (<85%), fragmentation, and low percentages (<25%) of REM sleep. With further development, this system may allow routine comprehensive characterization of sleep in ICU patients, previously out of reach due to the aforementioned constraints(Trompeo et al., 2011; Broughton and Baron, 1978).

## 5.3 Future work for improving `SLEEPNET`

Future work in the sleep medicine field includes several avenues. First, the algorithm can provide a pre-score to allow manual editing by the technologist to focus on low-confidence epochs, and thereby save time and increase the accuracy of staging. Second, with further validation the algorithm could replace manual scoring altogether in low-resource settings. Third, the algorithm could augment the accuracy of at-home monitors based on limited channels (frontal), which are increasingly available as alternatives to laboratory PSG.

Future development of SLEEPNET for use in the ICU environment could advance the care of patients with critical illness. In particular, sleep deprivation is believed to be a major preventable contributing factor to delirium, a condition of confusion and brain dysfunction that often leaves patients with enduring cognitive deficits even after recovery from the primary medical illness (Pandharipande et al., 2013). The ability to routinely monitor sleep in an accurate and automated manner is key to developing effective strategies to prevent and foreshorten delirium.

# 6. Conclusion

Using the largest archive assembled to date of real-world overnight sleep recordings, we have developed an automated sleep stage annotation system `SLEEPNET`, which trained a deep neural network to automatically label sequential EEG epochs with the 5 conventional sleep stages used in sleep medicine. We evaluated many different combinations of feature representation and models. On 1000 held-out testing patients, the best performing algorithm achieved an expert-algorithm level of inter-rater agreement of 85.76% with Kappa value 79.46%, exceeding previously reported levels of expert-expert inter-rater agreement for sleep EEG staging.

We have also developed a framework for model deployment to tackle inconsistent software configuration environments and the ease of use at the point of care in clinics. Our experience to date with the deployed model suggests that with further development `SLEEPNET` has great potential to be smoothly incorporated into real-world clinical workflows in both traditional sleep laboratories and in neurocritical care settings. Additionally, by automating a task previously requiring human expertise that is in short supply, `SLEEPNET` opens the way for extending the reach of sleep medicine beyond its current bounds.